\pdfoutput=1

\documentclass[11pt]{article}

\usepackage{acl}

\usepackage{times}
\usepackage{latexsym}

\usepackage[T1]{fontenc}

\usepackage{etoolbox}
\patchcmd{\quote}{\rightmargin}{\leftmargin 0.8em \rightmargin}{}{}

\usepackage[utf8]{inputenc}

\usepackage{microtype}

\usepackage{graphicx}
\usepackage{soul}
\usepackage{subcaption}
\usepackage{booktabs}
\usepackage{amsmath}
\usepackage{makecell}
\usepackage{array}
\usepackage{mdframed}
\usepackage{algorithmic}
\usepackage{amssymb}
\usepackage[ruled,vlined,linesnumbered]{algorithm2e}
\usepackage{multirow} 
\usepackage{longtable}
\usepackage{enumitem}
\usepackage[htt]{hyphenat}

\usepackage{silence}
\newcommand{\mae}{\ensuremath{\text{MAE}_{\text{test}}}}

\newcommand{\xtest}{x_{\text{test}}}

\newcommand{\xtrain}{x_{\text{train}}}

\newcommand{\xquery}{x_{\text{query}}}

\usepackage{kky}
\definecolor{violet}{RGB}{112,112,225}
\definecolor{purple}{RGB}{112,48,160}
\definecolor{green}{RGB}{0,176,80}
\definecolor{darkgreen}{RGB}{56,87,35}

\definecolor{darkblue}{RGB}{68,114,196}
\definecolor{darkred}{RGB}{192,0,0}
\definecolor{lightblue}{RGB}{51,204,255}
\definecolor{lightred}{RGB}{251,51,51}
\definecolor{grey}{RGB}{128,128,128}
\newcommand{\Scref}[1]{\S\ref{#1}}

\newcolumntype{P}[1]{>{\centering\arraybackslash}p{#1}}

\title{Beyond Demographics: Aligning Role-playing LLM-based Agents \\ Using Human Belief Networks}

\author{
    Yun-Shiuan Chuang \hspace{0.5cm}
    Krirk Nirunwiroj\textsuperscript{\textdagger} \hspace{0.5cm}
    Zach Studdiford\textsuperscript{\textdagger} \hspace{0.5cm}
    Agam Goyal \hspace{0.5cm} \\ 
    \textbf{Vincent V. Frigo} \hspace{0.5cm}
    \textbf{Sijia Yang} \hspace{0.5cm} 
    \textbf{Dhavan Shah} \hspace{0.5cm} 
    \textbf{Junjie Hu} \hspace{0.5cm} 
    \textbf{Timothy T. Rogers}\\
  University of Wisconsin-Madison\\
  \texttt{\{yunshiuan.chuang, nirunwiroj, studdiford, agoyal25\}@wisc.edu}\\
  \texttt{\{vfrigo, syang84, dshah, junjie.hu, ttrogers\}@wisc.edu}
}

\begin{document}
\maketitle
\def\thefootnote{\textdagger}\footnotetext{Joint second authors.}\def\thefootnote{\arabic{footnote}}

\begin{abstract}
Creating human-like large language model (LLM) agents is crucial for faithful social simulation. Having LLMs role-play based on demographic information sometimes improves human likeness but often does not. This study assessed whether LLM alignment with human behavior can be improved by integrating information from empirically-derived human belief networks. Using data from a human survey, we estimated a belief network encompassing 64 topics loading on nine non-overlapping latent factors. We then seeded LLM-based agents with an opinion on one topic, and assessed the alignment of its expressed opinions on remaining test topics with corresponding human data. Role-playing based on demographic information alone did not align LLM and human opinions, but seeding the agent with a single belief greatly improved alignment for topics related in the belief network, and not for topics outside the network. These results suggest a novel path for human-LLM belief alignment in work seeking to simulate and understand patterns of belief distributions in society. 
\end{abstract}


\section{Introduction}

With rapid advances in large language models (LLMs), there has grown increasing interest in using LLMs to simulate and understand dynamics of human communication and persuasion \cite{park2023generative,park2022social,chuang2024simulating,taubenfeld2024systematic}. Current LLMs can be prompted to role-play as individuals with particular demographic traits, sometimes then producing patterns of behavior that seem remarkably human-like. For instance, when asked to report the US unemployment rate when President Obama left office, ChatGPT will provide the exact answer; but if first instructed to role-play as a typical Democrat or Republican and asked the same question, the model produces incorrect, inflated estimates that mirror patterns of partisan bias in analogous human studies \cite{chuang2024wisdom}. Such results raise the possibility that, with strategic prompting, LLMs may serve as useful proxies for capturing the beliefs and attitudes of various socio-demographic groups.

\begin{figure}[tb!] 
\centering
\includegraphics[width=1\linewidth]{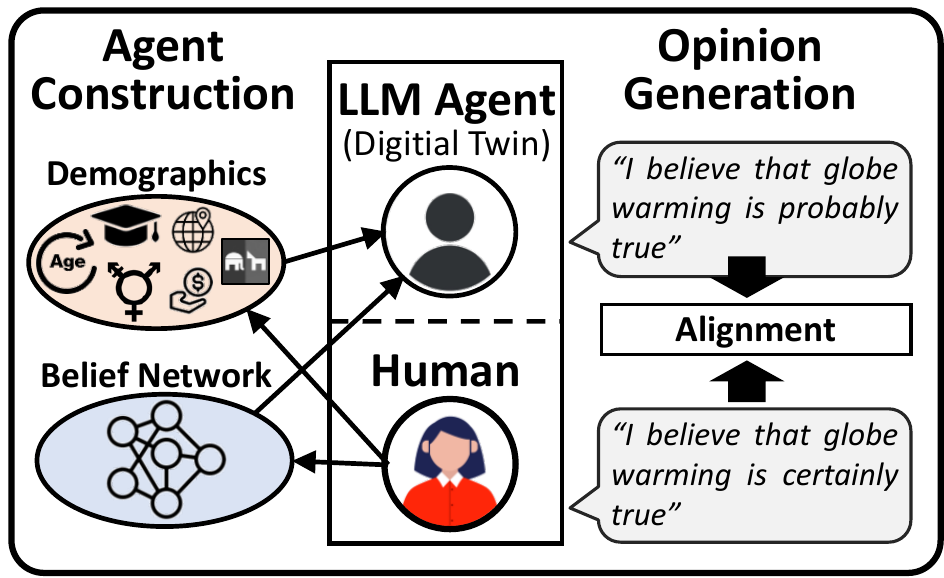}
\vspace{-2mm}
\caption{An LLM agent $i'$ is constructed as the ``digital twin'' of a human respondent $i$, based on their demographic information and belief network estimated from a belief survey. We then evaluate the alignment between the opinions generated by the agent ($o_{i'}$) and those expressed by the corresponding human respondent ($o_{i}$).} 
\label{fig:agent_vs_human}
\vspace{-4mm}
\end{figure}

Other recent work suggests, however, that the alignment between beliefs expressed by role-playing LLMs and matched human participants is unreliable at best. For instance, \citet{santurkar2023whose} found that LLMs tuned via human feedback generally reflect opinions from liberal and well-educated demographics and that having LLMs role-play as humans with different socio-demographic traits does not remediate this tendency. Similarly, \citet{sun2024random} had LLMs offer opinions on controversial issues while role-playing as humans with varying demographic characteristics, and found that the model only reflected corresponding human opinions on one of the ten total topics. \citet{chuang2024simulating} additionally found that, even when seeded with prompts specifying an initial belief that runs contrary to social consensus (e.g., ``global warming is a hoax''), LLMs quickly revert to the accepted ground-truth attitude after repeated interactions with other agents. Overall, this work suggests that LLMs fine-tuned with human feedback tend to adopt a consistent stance regardless of the demographic background they role-play---a behavior that may aid LLM fairness and value alignment, but limits their utility as models of human communicative dynamics.


This paper considers an alternative approach to aligning the attitudes expressed by role-playing LLMs and the human groups they are intended to emulate. The central idea relies on behavioral studies of human {\it belief networks}: the empirical observation that beliefs on different topics are not distributed at random across the population, but tend to cohere together in patterns of high-order covariation \cite{boutyline2017belief,vlasceanu2024network,keating2023persuasive,turner2022belief}. For instance, people who believe that government should support social welfare programs are also more likely to believe in higher taxes on the wealthy, strong union protections, and universal health care. Thus, knowing a person's opinion on one topic can carry rich information about their likely views on many others.
Because LLMs learn from vast amounts of human-generated language data, the weights they acquire and hence patterns of behaviors they exhibit may implicitly capture the tendency for various beliefs to co-occur in human populations, providing novel leverage for alignment. Specifically, human-LLM alignment may be guided, not just by socio-demographic role-playing, but also by instructing LLMs to hold a specific opinion on a representative topic. 

\begin{figure*}[th!] 
\centering
\includegraphics[width=1\linewidth]{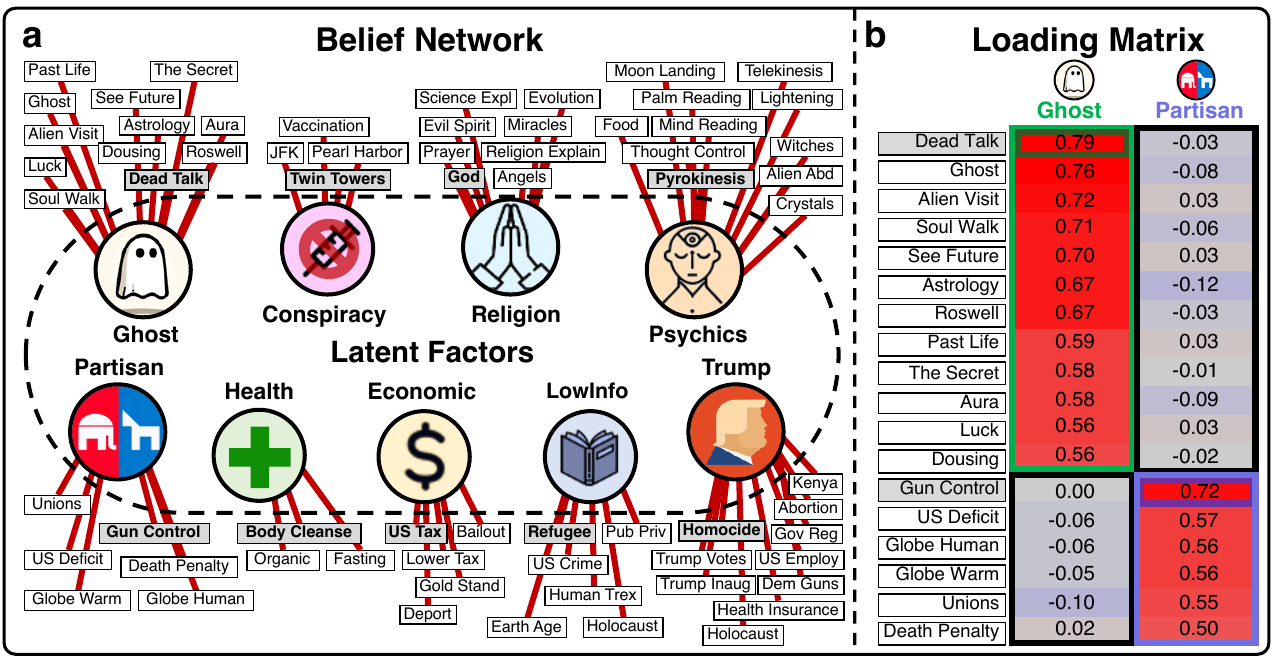}
\vspace{-2mm}
\caption{(a) The belief networks estimated by factor analysis from human respondents' responses on the Controversial Beliefs Survey. The nine central nodes are the orthogonal latent factors, and the leaves (rectangles) are the 64 individual topics $x$. The training topics $\xtrain$ are highlighted with grey backgrounds. (b) Factor loading matrix between two latent factors and their topics. Figure~\ref{fig:fa_loading_matrix} shows the full factor loading matrix and Table~\ref{tab:list_topic} the full statement of the each topic.}
\label{fig:belief_network_all_factors}
\end{figure*}

To test this idea,  
we considered a simple belief network constructed in prior work by applying factor analysis to a dataset measuring human beliefs across a diverse array of topics \cite{frigo2022examination}. Factor analysis decomposes patterns of covariation among expressed beliefs, identifying relationships between the beliefs themselves and a set of underlying latent factors. From this analysis we identified nine \textit{orthogonal} factors, each receiving high factor loadings (having strong associations) from several controversial beliefs. Each latent factor is associated with a distinct set of beliefs, with very little overlap between the beliefs linked to different factors. In other words, beliefs form distinct clusters with clear separation between clusters. Two example factors included a {\it ghost factor} grouping beliefs in various supernatural phenomena (e.g., talking to the dead) and a {\it partisan factor} grouping beliefs that are typically politically polarizing in the US (e.g., effectiveness of gun control).  We then considered how well the opinions of contemporary LLMs align with human participants when they role-play (a) without demographic information, (b) with demographic information only, or (c) with demographic information plus a belief on a specific topic that strongly aligns with either the same latent factor or a different latent factor. 
When seeding each model with such a belief, we additionally compared the effects of in-context learning (i.e. prompting) versus supervised fine-tuning.
The results suggest that attention to empirically-derived human belief networks provides a useful strategy for human-LLM alignment, moreso than demographic role-playing. 




\begin{figure*}[th!] 
\centering
\includegraphics[width=1\linewidth]{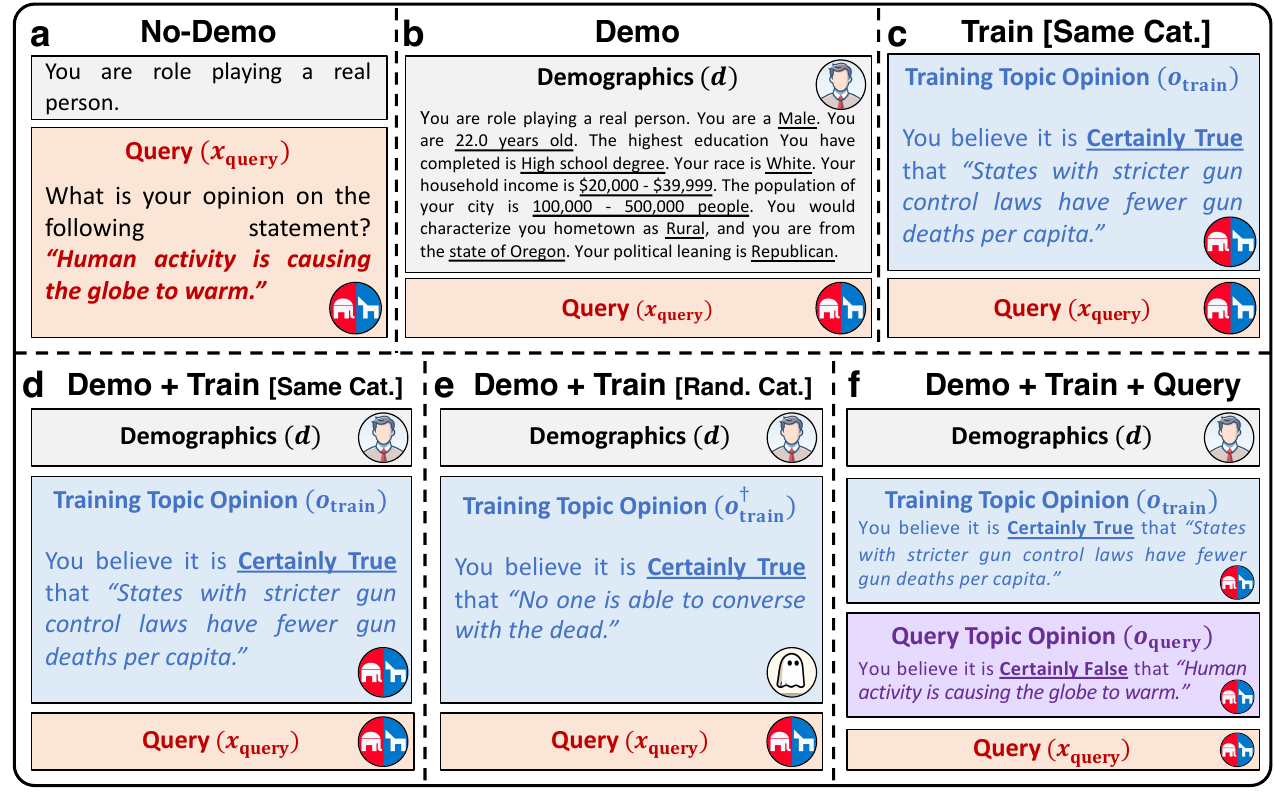}
\vspace{-2mm}
\caption{LLM agent construction conditions with different levels of respondent's information. (a) ``No-Demon'' baseline condition where the LLM role-plays without demographic information and we directly query the LLM about its opinion on the \textcolor{darkred}{query topic ($x_{\text{query}}$)}. (b) ``Demo'' baseline condition with demographic information ($d$). (c) ``Train [same category]'' baseline condition \textcolor{darkblue}{training topic opinion ($o_{\text{train}}$ on $x_{\text{train}}$)} from the same topic category as the query topic (in this example, they both belong to the ``Partisan'' category). (d) ``Demo+Train [same category]'' condition with demographic information plus \textcolor{darkblue}{training topic opinion ($o_{\text{train}}$ on $x_{\text{train}}$)} from the same topic category as the query topic. (e) ``Demo+Train [random category]'' baseline condition with demographic information, along with training topic opinion from a randomly selected topic category other than the query topic ($o_{\text{train}}^{\dagger}$ on $x_{\text{train}}^{\dagger}$) (in this example, the training topic is from the ``Ghost'' category). (f) ``Demo+Train+Query'' as a upper bound coundition with both training topic opinion (from the same category) and the \textcolor{purple}{query topic opinion ($o_{\text{query}}$ on $x_{\text{query}}$)}.} 
\label{fig:condition_agent_condition}
\vspace{-4mm}
\end{figure*}

\section{Preliminaries: LLM Agents as Human Digital Twins}

As depicted in Figure~\ref{fig:agent_vs_human}, we aim to construct an LLM agent $i'$ as the $i$-th human's ``digital twin'', such that their opinions $o$ on various topics $x$ are aligned. We first use information about human $i$ (e.g., their demographic information $d$) to create the corresponding LLM agent $i'$, and then query the agent's opinion ($o_{i'}$) on a wide range of topics. We then evaluate the human-LLM alignment by measuring the discrepancy between the actual human opinion $o_i$ and the LLM agent's opinion $o_{i'}$. Note that we use the term LLM-based ``\textit{agent}'' to refer to the digital twin because the instructed LLM is intended to produce behaviors that emulate the human individual they role-play \cite{park2023generative, shao2023character, zhou2023sotopia}.



\section{Methods}\label{sec:methods}

\subsection{Controversial Beliefs Survey} 


The specific opinions we assessed were taken from the {\it Controversial Beliefs Survey} developed in \citet{frigo2022examination}. The survey measures the direction and strength of belief across 64 topics spanning broad aspects of human knowledge, including history, science, health, religion, the supernatural, economics, politics, and conspiracy theories (see Table~\ref{tab:list_topic} in \Scref{app:list_topic} for the full list of topics). Topics were selected to elicit diverse opinions about their truthfulness (hence ``controversial beliefs''). Each belief was stated as a factual proposition (e.g., ``States with stricter gun control laws have fewer gun deaths per capita''), and participants rated their views about the truth of the statement on a six-point Likert scale ranging from ``Certainly false'' to ``Certainly true.'' 
Responses with high numbers indicate agreement with the rational/consensus ground truth. The dataset also has extensive demographic data from respondents, including age, gender, education level, household income, urban versus rural living environment, state of residence, and political leaning. 



The dataset includes ratings for $N=564$ individuals living in the US, collected from Amazon Mechanical Turk in 2018.\footnote{\url{https://mturk.com/}}. Formally, we denote the set of 64 topics as $\Xcal=\{x_j\}_{j=1}^M$ ($M=64$). The survey dataset $\Dcal=\{(d_i, x, o_i)| x\in\Xcal\}_{i=1}^N$ consists of the opinion responses from $N$ individuals, where the $i$-th individual having the demographic information $d_i$ expresses an opinion $o_i$ to the topic $x$. 
The respondents provide their opinions ($-3 \leq o_i \leq 3, o_i \neq 0$) for each statement on a 6-point Likert scale with the values $-3$: Certainly false, $-2$: Probably false, $-1$: Lean false, $+1$: Lean true, $+2$: Probably true, $+3$: Certainly true. No neutral value was provided so participants must minimally lean in one direction or the other.
The demographic and opinion data together were used to construct and evaluate the LLM agents (\Scref{sec:llm_agent_construction}). The survey dataset can be obtained by contacting its authors \cite{frigo2022examination}.


\subsection{Constructing a Belief Network using Factor Analysis}

Our objective was to find independent ``belief networks''---that is, groups of topics where expressed beliefs covaried across participants within each group but were independent between groups. To this end, we relied on a previous factor analysis \cite{frigo2022examination} that first computed a matrix of correlations in the ratings produced across participants for each pair of topics, then decomposed the resulting matrix into a set of orthogonal latent factors using principal component analysis (PCA) with Varimax rotation \citet{kaiser1958varimax}. 
The PCA yielded a factor {\it loading matrix} that encodes the loading (i.e., the association) between each topic and each latent factor. Nine latent factors were extracted based on the factor scree plot (\citealp{cattell1966scree}, see \Scref{app:fa_number_factors}), which together accounted for 72\% of the variance in the correlation matrix.
The belief network surrounding these nine factors are shown in Figure~\ref{fig:belief_network_all_factors}. For example, the {\it ghost factor} receives high loadings from 12 topics, all pertaining to supernatural or otherworldly beliefs; the {\it partisan factor} receives high loadings from 6 topics on highly polarized political issues. We referred to these topics as either belonging to the \textit{ghost topic category} or \textit{partisan topic category}, respectively. Hence, the nine orthogonal latent factors resulted in nine distinct \textit{topic categories}.
We took these 64 topics and the corresponding nine latent factors as the targets for our analysis of LLM alignment. The full factor analysis results, including the full factor loading matrix of the nine factors, are reported in \Scref{app:full_factor_loading_matrix}.


\subsection{LLM Agent Construction}
\label{sec:llm_agent_construction}

For each of the nine topic categories, we designated the topic possessing the highest loading as the model {\it training topic} ($x_{\text{train}}$). For each digital twin (role-playing LLM agent), the corresponding human opinion on the training topic ($o_{\text{train}}$) was used to customize the LLM agent (either through in-context learning or supervised fine-tuning, see below). Human opinions on the remaining 55 testing topics $x_{\text{test}}$ were not provided to the LLM agent; instead, the agent's expressed opinions $o_{\text{test}}$ on these topics were used to evaluate their alignment with the human respondents. We hypothesized that specifying the agent's opinion on the training topic might elicit a shared representation that generalizes to testing topics close within the belief network (i.e., sharing the same latent factor), but not those from the other belief network. 

For each human respondent $i$, we constructed an LLM agent $i'$ as their ``digital twin,'' using a set of strategies described below. For each twin created under a given strategy, we queried the LLM agent for its opinions on the training and test topics ($x_{\text{query}}$), and measured how ratings generated by the digital twins correlate with the true opinions expressed by corresponding human respondents. We then assessed how this measure of human-LLM belief alignment varied with different strategies for constructing the digital twin.

\paragraph{In-context Learning (ICL).} 
As shown in Figure~\ref{fig:condition_agent_condition}, these strategies involve initializing agents via in-context learning (ICL), with different information included in their \textit{system message} (see \Scref{sec:config_llm} and Appendix~\Scref{app:list_prompt_icl} for the prompts). 

\begin{enumerate}[label=\alph*.,leftmargin=15pt]\itemsep-0.3em
    \item \textbf{Baseline: No-Demo}. An LLM agent is role-playing a generic person without specific information about the human respondent (system message = ``\texttt{You are role playing a real person.}''). This provides a performance floor since there is no way for the LLM to align with a corresponding human participant. 
    \item \textbf{Baseline: Demo.} An LLM agent is constructed to role-play the $i$-th respondent by adding only the demographic information ($d_i$) in the prompt.
    \item \textbf{Baseline: Train [same category].} An LLM agent is constructed to role-play the $i$-th respondent by only adding the respondent's Likert-scale opinion on the training topic ($x_{\text{train}}$, $o_{\text{train}}$) and is assessed on other topics from the same topic category ($x_{\text{query}})$ within the belief network.
    \item \textbf{Demo+Train [same category].} In addition to demographic information, the LLM receives a respondent's Likert-scale opinion on the training topic ($x_{\text{train}}$, $o_{\text{train}}$) and is assessed on other topics from the same topic category ($x_{\text{query}})$ within the belief network. This is the critical condition of interest. 
    \item \textbf{Baseline: Demo+Train [random category].} This baseline condition is similar to Demo+Train [same category], 
    but the training topic opinion ($x_{\text{train}}^{\dagger}$, $o_{\text{train}}^{\dagger}$) belongs to a randomly selected topic category that is different from the query topic. This baseline allows us to determine whether adding respondent's Likert-scale opinion is only helpful when it belongs to the same belief network as the query topic ($x_{\text{query}}$).
    \item \textbf{Upper Bound: Demo+Train+Query.} This condition provides the human opinion rating on both the training topic ($x_{\text{train}}$, $o_{\text{train}}$) and the query topic ($x_{\text{query}}$, $o_{\text{query}}$) during the agent construction, providing an upper bound on generalization behavior. 
\end{enumerate}

\paragraph{Supervised Fine-tuning (SFT).} We also investigated whether seeding initial beliefs via supervised fine-tuning (SFT) can increase human-LLM alignment. Specifically, the correspondence between the demographic information $d$ and the corresponding opinion $o$ (on topic $x$) was used to fine-tune model weights via supervised learning, following analogous strategies to the in-context learning approaches described above. For example, for \textbf{Demo+Train [same category]}, we first construct the dataset $\Dcal_{\text{SFT}}=\{(d_i,x_{\text{train},i}),o_{\text{train},i}\}_{i=1}^N$ for each topic category. We then fine-tuned the LLM with input context providing the demographic information along with the training topic statement $(d,x_{\text{train}})$, and using the corresponding human Likert-scale response $o_{\text{train}}$ as the ground-truth output. After fine-tuning, we assessed the LLM agent's opinion on query topics $x_{\text{query}}$ belonging to the same topic category $x_{\text{train}}$ \footnote{For example, we fine-tuned an LLM on the respondents' opinions on the training topic for the Ghost topic category, then queried its opinion on the test topics in the Ghost topic category.}. Likewise, for \textbf{Baseline: Demo+Train [random category]}, it is similar to Demo+Train [same category] condition, but the training topic opinion ($x_{\text{train}}^{\dagger}$, $o_{\text{train}}^{\dagger}$) is from a different topic category as the query topic $\xquery$. Details of the fine-tuning procedure and the corresponding prompts are in \Scref{app:list_prompt_sft} and \Scref{app:fine_tuning}.

\section{Experimental Settings}

\begin{table*}[tbh!]
\centering
\Large
\resizebox{\linewidth}{!}{
\begin{tabular}{@{}p{2.5cm}lcccccccccc@{}} 
    \toprule
    \multirow{3}{*}{Model} & \multirow{3}{*}{Condition} & \multicolumn{9}{c}{Topic Categories} & \multirow{3}{*}{Average} \\ 
    \cmidrule(l){3-11} 
    & & Ghost & Psychics & Religion & Trump & Partisan & Economic & LowIndo & Health & Conspiracy & \\
    \midrule
    \multirow{7}{*}{\shortstack[l]{ChatGPT}}     
    & \textit{Baselines} & & & & & & & & & & \\
    & \hspace{1em} No-Demo & 2.33 & 2.26 & 1.81 & 1.17 & 1.43 & 1.42 & 1.29 & 1.62 & 1.80 & 1.68 \\
    & \hspace{1em} Demo & 2.58 & 2.28 & 1.87 & 1.23 & 1.41 & 1.51 & 1.21 & 1.66 & 1.51 & 1.70 \\
    & \hspace{1em} Train [Same Cat.] & 1.48 & 1.46 & 1.80 & 1.18 & 1.36 & 1.48 & 1.23 & 1.60 & 1.76 & 1.48 \\
    & \hspace{1em} Demo + Train [Rand. Cat.] & 2.26 & 1.86 & 1.93 & 1.29 & 1.49 & 1.63 & 1.26 & 1.80 & 1.53 & 1.67 \\    
    & \textbf{Demo + Train [Same Cat.]} & \textbf{1.26} & \textbf{1.27} & \textbf{1.72} & \textbf{1.14} & \textbf{1.34} & \textbf{1.23} & \textbf{1.15} & \textbf{1.53} & \textbf{1.40} & \textbf{1.34} \\
    & \textit{Upper Bound} & & & & & & & & & & \\
    & \hspace{1em} Demo + Same Train + Query & 0.41 & 0.48 & 0.30 & 0.63 & 0.28 & 0.09 & 0.82 & 0.30 & 0.46 & 0.42 \\
    \cmidrule(r){2-12}
    & Relative Gain (\%) $\uparrow$ & 60.83 & 56.11 & 9.55 & 15.00 & 6.19 & 19.72 & 15.38 & 9.56 & 10.48 & 22.54 \\
    \midrule
    \multirow{7}{*}{\shortstack[l]{GPT-4o mini} }   
    & \textit{Baselines} & & & & & & & & & & \\
    & \hspace{1em} No-Demo & 1.49 & 1.33 & 1.90 & 1.21 & 1.19 & 1.30 & 1.31 & 2.03 & 1.40 & 1.46 \\
    & \hspace{1em} Demo & 1.46 & 1.21 & 1.68 & 1.17 & 1.19 & 1.24 & 1.23 & 1.41 & 1.42 & 1.33 \\
    & \hspace{1em} Train [Same Cat.] & 1.05 & 0.96 & 1.36 & 1.06 & 1.18 & 1.19 & 1.21 & 1.42 & 1.32 & 1.19 \\
    & \hspace{1em} Demo + Train [Rand. Cat.] & 1.44 & 1.23 & 1.53 & 1.28 & 1.24 & 1.22 & 1.19 & 1.58 & 1.41 & 1.35 \\    
    & \textbf{Demo + Train [Same Cat.]} & \textbf{1.00} & \textbf{0.96} & \textbf{1.31} & \textbf{1.06} & \textbf{1.15} & \textbf{1.19} & \textbf{1.16} & \textbf{1.37} & \textbf{1.28} & \textbf{1.16} \\
    & \textit{Upper Bound} & & & & & & & & & & \\
    & \hspace{1em} Demo + Same Train + Query & 0.04 & 0.05 & 0.03 & 0.64 & 0.01 & 0.02 & 0.14 & 0.04 & 0.14 & 0.12 \\
    \cmidrule(r){2-12}
    & Relative Gain (\%) $\uparrow$ & 32.39 & 21.55 & 22.42 & 20.75 & 3.39 & 4.10 & 6.42 & 2.92 & 10.94 & 13.88 \\
    \midrule
    \multirow{7}{*}{\shortstack[l]{Mistral}}     
    & \textit{Baselines} & & & & & & & & & & \\
    & \hspace{1em} No-Demo & 1.75 & 1.63 & 1.64 & 1.33 & 1.20 & 1.07 & 1.49 & 1.30 & 1.44 & 1.43 \\
    & \hspace{1em} Demo & 1.82 & 1.93 & 1.68 & 1.49 & 1.27 & 1.16 & 1.49 & 1.39 & 1.38 & 1.51 \\
    & \hspace{1em} Train [Same Cat.] & 1.46 & 1.02 & 1.46 & 1.46 & 1.25 & 1.12 & 1.44 & 1.44 & 1.28 & 1.33 \\
    & \hspace{1em} Demo + Train [Rand. Cat.] & 1.93 & 1.79 & 1.60 & 1.56 & 1.35 & 1.22 & 1.70 & 1.36 & 1.45 & 1.55 \\    
    & \textbf{Demo + Train [Same Cat.]} & \textbf{1.36} & \textbf{1.71} & \textbf{1.41} & \textbf{1.05} & \textbf{1.25} & \textbf{1.12} & \textbf{1.12} & \textbf{1.32} & \textbf{1.27} & \textbf{1.29} \\
    & \textit{Upper Bound} & & & & & & & & & & \\
    & \hspace{1em} Demo + Same Train + Query & 0.71 & 0.39 & 0.86 & 0.77 & 0.59 & 0.55 & 0.65 & 1.04 & 0.55 & 0.68 \\
    \cmidrule(r){2-12}
    & Relative Gain (\%) $\uparrow$ & 41.44 & 14.29 & 32.93 & 61.11 & 2.94 & 6.56 & 44.05 & 20.00 & 13.25 & 26.29 \\
    \midrule
    \multirow{7}{*}{\shortstack[l]{LLaMA 3.1}}     
    & \textit{Baselines} & & & & & & & & & & \\
    & \hspace{1em} No-Demo & 2.55 & 2.40 & 1.88 & 1.86 & 2.04 & 2.54 & 1.52 & 1.54 & 2.11 & 2.05 \\
    & \hspace{1em} Demo & 2.36 & 2.42 & 1.85 & 1.50 & 1.45 & 2.33 & 1.47 & 1.50 & 2.35 & 1.91 \\
    & \hspace{1em} Train [Same Cat.] & 2.21 & 2.28 & 1.82 & 1.44 & 1.63 & 1.86 & 1.48 & 1.63 & 2.77 & 1.90 \\
    & \hspace{1em} Demo + Train [Rand. Cat.] & 2.70 & 2.64 & 2.03 & 1.69 & 1.87 & 2.48 & 1.80 & 1.97 & 2.28 & 2.16 \\    
    & \textbf{Demo + Train [Same Cat.]} & \textbf{2.07} & \textbf{1.88} & \textbf{1.81} & \textbf{1.19} & \textbf{1.32} & \textbf{1.69} & \textbf{1.35} & \textbf{1.07} & \textbf{2.00} & \textbf{1.60} \\
    & \textit{Upper Bound} & & & & & & & & & & \\
    & \hspace{1em} Demo + Same Train + Query & 1.76 & 1.04 & 1.42 & 0.96 & 0.56 & 1.47 & 0.72 & 0.96 & 0.65 & 1.06 \\
    \cmidrule(r){2-12}
    & Relative Gain (\%) $\uparrow$ & 48.33 & 39.13 & 9.30 & 57.41 & 14.61 & 74.42 & 16.00 & 79.63 & 21.05 & 39.99 \\
    \bottomrule
\end{tabular}
}
\caption{Mean absolute error ($\mae$) between human respondents and the corresponding LLM agents for each topic category across various LLM agent construction conditions through in-context learning (ICL). The bottom row presents the relative gain (\%) as the percentage improvement from the Demo Baseline to the Upper Bound condition for the Demo + Train [Same Cat.] condition. The lower the $\mae$ and higher the relative gain, the higher the human-LLM alignment. The condition of our main interest (i.e., Demo + Train [Same Cat.] condition) is boldfaced, which also has the best alignment.}
\label{tab:result_llm_vs_human_all_models}
\end{table*}

\label{sec:exp_settings}

\subsection{Configuration for LLM Agents}
\label{sec:config_llm}

We evaluated the LLM agents using the following models: ChatGPT (\texttt{gpt-3.5-turbo-0125}; \citealp{openaiIntroducingChatGPT}), GPT-4o mini (\texttt{gpt-4o-mini-2024-07-18}), Mistral (\texttt{Mistral-7B-Instruct-v0.2}; \citealp{jiang2023mistral}), and LLaMA 3.1 (\texttt{Llama-3.1-8B-Instruct}; \citealp{touvron2023llama}), all with temperature of $0.7$. In sensitivity analyses, we consider other temperature values $T \in \{0, 1\}$. During initialization, the demographic background was incorporated into the model's \textit{``system messages.''} The opinion queries ($\xquery$) were fed to the agent through the model's \textit{``user messages.''} When using in-context learning (\Scref{sec:llm_agent_construction}), the training/query topic opinions were also included in the model's \textit{``system messages.''} The LLM agents were constructed through LangChain \cite{langchain}. For our compute resources, see \Scref{app:compute_resources}.

\subsection{Evaluation Metrics}
To evaluate the ``human-likeness'' of the LLM agents' opinions, for each topic category in the survey, we computed the mean absolute error ($\mae$) between the human opinion ($o_{i}$) and that generated by the twinned LLM agent ($o_{i'}$) across the testing topics ($\xtest$). Formally, $\mae=\frac{1}{|\Xcal_{\text{test}}|}\sum_{x\sim\Xcal_{\text{test}}}{|o_{i,x}-o_{i',x}|}$, which is the mean discrepancy between the opinions of human respondents and LLM agents across all test topics ($\Xcal_{\text{test}}$) within the topic category. The metric $\mae$ ranges from 0 to 4, where 0 indicates perfect agreement and 4 is the maximum possible disagreement. In addition, because we are interested in the additional value of belief network beyond demographic information, we calculate Relative Gain (\%) as the percentage improvement from the Demo Baseline to the Upper Bound condition for the Demo + Train [same category] condition, i.e., Relative Grain (\%) = ($\mae$ of ``Baseline: Demo.'' $-$ $\mae$ of ``Demo+Train [same category]'') /  ($\mae$ of ``Baseline: Demo.'' $-$ $\mae$ of ``Upper Bound: Demo+Train+Query'') $\times 100$ (\%). The Relative Gain is $0 \%$ if belief network provides no additional benefit, and $100 \%$ if the inclusion of belief network boosts the alignment to the supervised upper bound.

\subsection{Supervised Fine-tuning (SFT)}

For LLM agents constructed through supervised fine-tuning (\Scref{sec:llm_agent_construction}), we used the ChatGPT model \texttt{gpt-3.5-turbo-0125}'s fine-tuning API. Critically, because the label (i.e., opinion response $o$) is usually not balanced in a given topic (e.g., more people believing that ghosts are real than those who don't), we upsampled the $o$ to ensure equal numbers of responses across the six Likert scale values. Pilot work found that, without upsampling, the fine-tuned LLM agent predominantly produced the most frequent opinion response $o_{\text{majority}}$ in $\Dcal_{\text{SFT}}$. 
Given that the primary aim of the SFT setting is to demonstrate the generalizability of our methods beyond the ICL framework, and recognizing that SFT is inherently more computationally demanding, we concentrate our investigation on two latent factors: the Ghost factor and the Partisan factor\Scref{app:fine_tuning} lists the hyperparameters for fine-tuning.


\section{Results}\label{sec:results}



\paragraph{Demographic information alone does not align the LLM agent's opinion.}

As shown in Table~\ref{tab:result_llm_vs_human_all_models}, incorporating solely the demographic information (the Demo condition) fails to align LLM agents with human respondents. The $\mae$ of the Demo condition is similar to the No-Demo condition, indicating that the demographic information alone does not help LLM agents align with the human respondents they role-play.


\paragraph{Specifying the agent's opinion on a training topic aligns other beliefs in the same network.}
When the LLM is instructed to adopt the twinned human's opinion on the training topic ($x_{\text{train}}$, $o_{\text{train}}$), its expressed opinions on other topics in the same belief network correlate significantly (i.e., become aligned) with the corresponding human opinions (Demo+Train [same category] condition; indicated by lower $\mae$). For example, when an LLM agent is initialized to believe that ``some people can communicate with the dead'' (the training topic $x_{\text{train}}$), then the LLM agent becomes more likely to also believe that ``people can project their soul out of their body'' (the query topic $x_{\text{query}}$). Concretely, when averaged across nine topic categories, the inclusion of the training topic opinion reduces $\mae$ from 1.70 (the Demo condition; ChatGPT) to 1.34 (the Demo+Train [same category] condition), representing a 22.54\% relative gain. Critically, this effect is limited to topics within the same belief network. If the training topic is from a different topic category (e.g., about the effectiveness of gun control law; Demo+Train [random category] baseline condition), the opinion of the LLM agent on the query topic remains unaligned with the corresponding human ($\mae = 1.67$). This supports our hypothesis -- opinions on one topic encourage the LLM agents to align their opinions only when the topics are adjacent in the belief network. 


%

\paragraph{Combining demographic information and training topic opinion reaches the best alignment.} While demographic information does not improve alignment on its own (the Demo condition), does it offer any benefit? The contrast between the Demo + Train [same category] condition and the Train [same category] baseline condition answers this question. When removing demographic information from the Demo + Train [same category] condition, the $\mae$ increases from 1.34 to 1.48 (ChatGPT), and the relative gain decreases from 22.54 \% to 17.19 \%. This shows that to reach the best alignment, both the training topic opinion and the demographic should be included.



\paragraph{Alignment does not reflect superficial repetition.}

\begin{table}[tb!]
\centering
\small
\resizebox{\linewidth}{!}{
    \begin{tabular}{@{}lrr@{}}
        \toprule
        \multirow{2}{*}{Model} & \multicolumn{2}{c}{Demo + Train [Same Cat.]} \\
        \cmidrule(l){2-3} 
        & [Original] & [Balanced] \\
        \midrule
        ChatGPT \\
        \hspace{1em} Average MAE\textsubscript{test} & 1.34 & 1.41 \\
        \hspace{1em} Average Relative Gain (\%) $\uparrow$& 22.54 & 22.19 \\
        GPT-4-o-mini \\
        \hspace{1em} Average MAE\textsubscript{test} & 1.16 & 1.21 \\
        \hspace{1em} Average Relative Gain (\%) $\uparrow$& 13.88 & 9.91 \\
        Mistral \\
        \hspace{1em} Average MAE\textsubscript{test} & 1.29 & 1.31 \\
        \hspace{1em} Average Relative Gain (\%) $\uparrow$& 26.29 & 24.67 \\
        LLaMA 3.1 \\
        \hspace{1em} Average MAE\textsubscript{test} & 1.60 & 1.71 \\
        \hspace{1em} Average Relative Gain (\%) $\uparrow$& 39.99 & 23.93 \\
        \bottomrule
    \end{tabular}
}
\caption{Average $\mae$ and average relative gain of the Demo+Train [Same Cat.] condition for the original condition (``[Original]'') and the variant where we balance the label distribution (``[Balanced]''). Note that balancing the label distribution does not change the superiority of Demo+Train [same category] condition when compared with the Demo condition.}
\label{tab:result_llm_vs_human_balanced_label}
\end{table}

Does increased alignment following the Demo+Train [same category] condition arise from a model tendency to simply repeat the opinion provided for the training topic? Such a pattern might appear to lead to increased alignment simply because the training topic opinion, by definition, correlates with opinions on other topics in the same belief network.
To address this concern, we conducted an additional experiment in which we balanced the label distribution in the prompting contexts by 
constructing reversed framing statements that entail the same semantic meaning. We then included both the original and reversed framing statements in the context. For example, for the original statement ``You believe it is \textit{\underline{certainly true}} that `States with stricter gun control laws have \textit{\underline{fewer}} gun deaths per capita''', the reversed frame stated ``You believe it is \textit{\underline{certainly false}} that `States with stricter gun control laws have \textit{\underline{more}} gun deaths per capita'''. Both statements were included in the context in random order so the LLM cannot show increased alignment by merely repeating the training topic opinion. Table~\ref{tab:result_llm_vs_human_balanced_label} shows that the LLMs continue to show significant alignment with human opinions
(low $\mae$) in this case, an effect that
must reflect the meaning of the joint information $(x_{\text{train}}, o_{\text{train}})$ rather than the opinion label $o_{\text{train}}$ alone.

\paragraph{Sensitivity Analyses} We evaluated the sensitivity of our result to randomness due to different temperature values when using temperature sampling. Across $T \in \{0, 0.7, 1\}$ using ChatGPT, the results showed consistent trends (Table~\ref{tab:result_llm_vs_human_chatgpt_sensitivity_temperature}).

\paragraph{Supervised fine-tuning yields similar results.}
As shown in Table~\ref{tab:result_llm_vs_human_fine_tune_chatgpt}, when the agents are fine-tuned with a training topic $x_{\text{train}}$, they also express more human-like opinions on query topics belonging to the same belief network (i.e., lower $\mae$; the Demo+Train [same category] condition), but not on those belonging to a different network (Demo+Train [random category] condition)--a pattern of results qualitatively similar to in-context learning.
\begin{table}[tb!]
\centering
\small
\resizebox{\linewidth}{!}{
    \begin{tabular}{@{}lrr@{}}
        \toprule
        Condition & \multicolumn{2}{c}{Topic Category} \\
        \cmidrule(l){2-3} 
        & Ghost & Partisan \\
        \midrule
        \textit{Baselines.} \\
        \hspace{1em} Demo & 2.58 & 1.41 \\
        \hspace{1em} Demo + Train [Rand. Cat.] & 2.31 & 1.35 \\
        \textbf{Demo + Train [Same Cat.]} & \textbf{1.29} & \textbf{1.25} \\
        \textit{Upper bound} \\
        \hspace{1em} Demo + Same Train + Truth & 0.41 & 0.28 \\
        \midrule
        Relative Gain (\%) & 59.45 & 14.16 \\
        \bottomrule
    \end{tabular}
}
\caption{Mean absolute error ($\mae$) between human respondents and the corresponding LLM agents for each topic category across various LLM agent construction conditions through supervised fine-tuning (SFT). The condition of our main interest (i.e., Demo + Train [Same Cat.] condition) is boldfaced, which also has the best alignment.}
\label{tab:result_llm_vs_human_fine_tune_chatgpt}
\end{table}

\section{Related Work}\label{sec:related_work}

\paragraph{Aligning human and LLM opinions.}
Recent studies highlight both the potential and the limitations of using LLMs to emulate human opinions \cite{argyle2023out,santurkar2023whose,sun2024random,feng2023pretraining,chuang2024simulating,chuang2024wisdom}. \citet{argyle2023out} showed that LLMs conditioned on demographic backstories can emulate human voting preferences and language use, but did not investigate topic-specific opinions. \citet{santurkar2023whose} found that different models have different inherent opinions that often align with liberal, high-income, well-educated demographics, and that 
these opinions could not be shifted by providing demographic role-playing information. 
The current paper replicates this finding, but additionally suggests that alignment may be shifted via belief networks. To the best of our knowledge no prior work has studied such effects.

\paragraph{Belief networks.}
A great deal of prior work has studied human belief networks \cite{boutyline2017belief,vlasceanu2024network,keating2023persuasive,turner2022belief, powell2023modeling,devine2015ideological,jewitt2016ideological,baldassarri2014neither,brandt2021evaluating} and has developed a range of approaches beyond factor analysis for characterizing these
including partial correlation networks \cite{turner2022belief} or Bayesian networks \cite{powell2023modeling}. 
Such networks have been shown to predict ``spillover effects'' of attitude changes across related topics \cite{turner2022belief,powell2023modeling} in human participants, where a change in a given topic can ripple through the belief network and influence related topics. In the present study, we investigated whether we can leverage the belief network derived from human data to construct LLM agents that more accurately reflect human opinions.

\section{Conclusion}

We investigated the use of empirically-derived belief networks for promoting alignment of expressed beliefs between Large Language Model (LLM) agents and twinned human participants. We showed that demographic role-playing alone does not produce significant alignment, but that initializing an agent with a human
opinion on one topic then aligns opinions on nearby topics within the belief network. The effect does not extend to distant topics within the network. We found similar effects for in-context learning and supervised fine-tuning, for both a proprietary and an open-source LLM. This work highlights a novel and potentially powerful means of enhancing LLM agents' alignment with human opinions.


\section*{Limitations}

\paragraph{The scope of topics}
We considered just 18 topics derived from two orthogonal latent factors identified in prior work. 
While the Partisan topics are of public interest and the Ghost topics explore an orthogonal dimension, future research could greatly the scope of topics.

\paragraph{The structure of the belief network.}
We considered belief networks based on two highly distinct clusters to facilitate evaluation. Other studies have used more sophisticated models, such as Bayesian networks \cite{powell2023modeling}, which allow for precise predictions about topic interrelations. Future work could apply such methods to better characterize belief networks.

\paragraph{The actions of the LLM agents.}
Our LLM agents expressed their opinions through Likert-scale ratings. This facilitated direct comparison with human responses but may not fully capture the expression of opinions in real-world settings like social media communication. Future studies could explore more complex actions (e.g., writing social media posts) to assess their human-likeness in realistic applications.

\section*{Ethics Statement}

We aim to develop LLM agents capable of simulating realistic human communicative dynamics, including the expression of potentially harmful beliefs such as misconception about the reality of global warming.  Our objective is 
to facilitate a deeper understanding of social phenomena like misinformation spread in order to 
identify strategies that mitigate these challenges effectively. Note that under the current setting, the LLM agents only produce Likert-scale ratings from a fixed set of options. Therefore, they are not able to produce unexpected harmful responses. We will release our code base solely for research purposes, and adhere to the terms of use by OpenAI's API \footnote{\url{https://openai.com/policies/terms-of-use}} and their MIT license \footnote{\url{https://github.com/openai/openai-openapi/blob/master/LICENSE}}, as well as Mistral AI's non-production license (MNPL) \footnote{\url{https://mistral.ai/licenses/MNPL-0.1.md}}.

\section*{Acknowledgements}

We thank the reviewers, the area chair for their feedback. This work was funded by the Multi University Research Initiative grant from the Department of Defense, W911NF2110317 (with Rogers as Co-I), Cohesive and Robust Human-Bot Cybersecurity Teams, the John S. and James L. Knight Foundation (Award Number: MSN231314), and the National Science Foundation through the Convergence Accelerator Track F: Course Correct: Precision Guidance Against Misinformation (Agency Tracking Number: 2230692; Award Number: MSN 266268).

\newpage
\bibliography{main}

\newpage
\pagebreak
\appendix

\newpage
\appendix
\label{sec:appendix}

\section{List of the 64 Topics in the Belief Survey}
\label{app:list_topic}
Table~\ref{tab:list_topic} shows the full stetements of the 64 topics in the Belief Survey, including the topic category to which they belong according to the factor analysis result, along with whether they belong to the training or the test partition.

\clearpage
\onecolumn
{
    \small
    \centering  
    \begin{longtable}{@{}llp{10cm}@{}}
        \toprule
        Topic Category &
          Topic  Name &
          Topic Statement
           \\
       \toprule
        Ghost &
          Dead Talk &
          No one is able to converse with the dead.
           \\
         &
          Ghost &
          After someone has died it is not possible to see his or her ghost. 
           \\
         &
          Alien Visit &
          Intelligent beings from outer space have not visited the Earth via spaceships. 
           \\
         &
          Soul Walk &
          It is not possible for anyone to project their soul out of their body. 
           \\
         &
          See Future &
          No one is capable of having visions that accurately predict future events. 
           \\
         &
          Astrology &
          The position of the planets at the time of your birth has no influence on your personality. 
           \\
         &
          Roswell &
          No alien spacecraft has ever crashed near Roswell, New Mexico. 
           \\
         &
          Past Life &
          Nobody can accurately remember living a past life. 
           \\
         &
          The Secret &
          Strongly visualizing your fondest wish does not make it more likely to become a reality. 
           \\
         &
          Aura &
          Health cannot be improved by manipulating a person's aura or electrical field. 
           \\
         &
          Luck &
          ``Lucky streaks''
          where random events are more likely to favor a person are not real.
          \\
         &
          Dousing &
          Nobody can sense water using only a forked stick. 
           \\
       \cmidrule{2-3}
        Psychics &
          Pyrokinesis &
          Nobody can start fires just by thinking about it. 
           \\
         &
          Thought Control &
          Nobody can control another's actions with their mind. 
           \\
         &
          Food &
          Food dropped on the ground for less than five seconds can become contaminated.
           \\
         &
          Palm Reading &
          It is not possible to predict future life events from markings on a person's palm. 
           \\
         &
          Telekinesis &
          No one is capable of moving objects with his or her mind. 
           \\
         &
          Witches &
          Witches cannot influence events by using magic. 
           \\
         &
          Mind Reading &
          No one is capable of reading another person's thoughts. 
           \\
         &
          Moon Landing &
          US astronauts have landed on the moon. 
           \\
         &
          Crystals &
          Crystals do not have unexplained powers. 
           \\
         &
          Lightening &
          Lightning can strike twice in the same place. 
           \\
         &
          Alien Abd &
          Human beings have not been abducted by aliens from outer space. 
           \\
       \cmidrule{2-3}       
        Religion &
          God &
          God does not exist.
           \\
         &
          Prayer &
          Prayer cannot cure illness. 
           \\
         &
          Angels &
          Angels are not real. 
           \\
         &
          Religion Explain &
          Religion does not provide the most accurate explanation for how the universe came into existence. 
           \\
         &
          Evil Spirit &
          It is not possible for a person's actions to be controlled by an evil spirit. 
           \\
         &
          Science Expl &
          Everything that happens can eventually be explained by science. 
           \\
         &
          Miracles &
          Miracles that defy the laws of nature cannot happen. 
           \\
         &
          Evolution &
          Species living on the Earth today have not always existed in their present form. 
           \\
      \cmidrule{2-3} 
      Trump   &
          Homicide &
          In the US, about 80\% of white homicide victims are killed by white people.  
           \\
         &
          Trump Inaug &
          More people attended the inauguration of Barack Obama than the inauguration of Donald Trump. 
           \\
         &
          Kenya &
          Barack Obama was born in Hawaii. 
           \\
         &
          US Employment &
          The US unemployment rate in 2016 was lower than 40\%. 
           \\
         &
          Gov Reg &
          Government regulations do not always stifle economic growth. 
           \\
         &
          Holocaust &
          The Nazi government in Germany murdered approximately 6 million Jewish people during the second world war. 
           \\
         &
          Trump Votes &
          Hilary Clinton received the most overall votes in the 2016 Presidential election. 
           \\
         &
          Abortion &
          Strongly Republican states have higher rates of abortion than strongly Democratic states. 
           \\
         &
          Dem Guns &
          The official platform of the Democratic Party does not seek to repeal the 2nd Amendment. 
           \\
         &
          Health Insurance &
          Since the Affordable Care Act (Obamacare) passed, more Americans have health insurance. 
           \\
       \cmidrule{2-3}            
        Partisan &
          Gun Control &
          States with stricter gun control laws have fewer gun deaths per capita.
           \\
         &
          US Deficit &
          The US deficit decreased after President Obama was elected.
           \\
         &
          Globe Human &
          Human activity is causing the globe to warm.
           \\
         &
          Globe Warm &
          The global climate is rapidly growing warmer.
           \\
         &
          Unions &
          States with strong union protections have lower unemployment than states without such protections.
           \\
         &
          Death Penalty &
          States that have the death penalty have higher rates of violent crime on average.
           \\
       \cmidrule{2-3}            
        Economic &
          US Tax &
          The United States doesn't have the highest federal income tax rate of any Western country.  
           \\
         &
          Deport &
          President G. W. Bush deported fewer undocumented immigrants than President Obama.  
           \\
         &
          Lower Tax &
          Lowering taxes does not always lead to economic growth.  
           \\
         &
          Bailout &
          The rescue of big banks by the federal government aided recovery from the 2008 recession.  
           \\
         &
          Gold Stand &
          Returning to the Gold Standard would make the US more vulnerable to a recession.  
           \\
       \cmidrule{2-3}        
        LowInfo &
          Refugee &
          In 2016 fewer than 100,000 refugees from the Middle East were granted permission to live in the United States.  
           \\
         &
          US Crime &
          The violent crime rate in the US has declined over the past 10 years.  
           \\
         &
          Earth Age &
          The Earth is not around 6,000 years old.  
           \\
         &
          Human Trex &
          The Tyrannosaurus Rex and humans did not live on the Earth at the same time.  
           \\
         &
          Pub Priv &
          For a given level of education, private-sector workers typically earn more than government workers.  
           \\
       \cmidrule{2-3}        
        Health &
          Body Cleanse &
          A ``body cleanse''
          in which you consume only particular kinds of nutrients over 1-3 days
          does not help your body to eliminate toxins.  \\
         &
          Organic &
          Organic foods are not healthier to eat than non-organic foods.  
           \\
         &
          Fasting &
          Regular fasting will not improve your health.  
           \\
       \cmidrule{2-3}        
        Conspiracy &
          Twin Towers &
          The twin towers were not brought down from the inside by explosives during the 9/11 attack.  
           \\
         &
          JFK &
          Only one gunman was involved in the assassination of John F. Kennedy.  
           \\
         &
          Pearl Harbor &
          President Roosevelt did not know about the attack on Pearl Harbor ahead of time.  
           \\
         &
          Vaccination &
          Vaccinations cannot cause Autism.   \\
          \bottomrule
    \caption{The statements of the 64 topics in the Belief Survey, including the topic category to which they belong according to the factor analysis result.}
    \label{tab:list_topic}      
    \end{longtable}
}
\clearpage
\twocolumn

\section{The Prompts for LLM Agent Construction Through In-context Learning (ICL)}
\label{app:list_prompt_icl}

Table~\ref{tab:list_prompt_icl} shows the prompts we use to construct and query the LLM agents in the in-context learning setting (\Scref{sec:llm_agent_construction}). Different LLM agent construction conditions include various sets of the prompt types. The parts enclosed in curly brackets ``$\{ \}$'' are the placeholders (e.g., $\{\text{demo\_age}\}$, $\{\text{\text{query}\_topic\_statement}\}$), where they are filled with actual information from either the respondents or the belief survey. As shown in Figure~\ref{fig:condition_agent_condition} and \Scref{sec:llm_agent_construction}, in the \textbf{Baseline: No-Demo} condition,  only the ``Query'' prompt is included. In the \textbf{Baseline: Demo} condition, both the prompt types ``Demographics'' and ``Query'' are included. In the \textbf{Demo + Train} conditions (both [same category] and [random category]), the prompt types include ``Demographics'', ``Training Topic Opinion'', and ``Query''. In the \textbf{Upper Bound: Demo + Train + Query} condition, the prompt types include ``Demographics'', ``Training Topic Opinion'', ``Query Topic Opinion'', and ``Query''.


\begin{table*}[th!]
\small
\centering
\resizebox{\linewidth}{!}{  
    \begin{tabular}{@{}p{2cm}p{2cm}p{7cm}p{7cm}@{}}
        \toprule
        Prompt Type & Message Type (LangChain) & Prompt Template & Example \\
        \toprule
        Demographics & \textit{System Message} &
        You are role playing a real person. You are a \{demo\_gender\}. You are \{demo\_age\} years old. The highest education You have completed is \{demo\_education\}. Your race is \{demo\_race\}. Your household income is \{demo\_income\}. The population of your city is \{demo\_city\_pop\}. You would characterize your hometown as \{demo\_urban\_rural\}, and you are from the state of \{demo\_state\}. Your political leaning is \{demo\_party\}.& 
        You are role playing a real person. You are a \{Male\}. You are \{41\} years old. The highest education You have completed is \{Some college but no degree\}. Your race is \{White\}. Your household income is \{$40,000 - $59,999\}. The population of your city is \{100,000 - 500,000\}. You would characterize your hometown as \{Urban (City)\}, and you are from the state of \{Florida\}. Your political leaning is \{Democrat\}. \\
        \midrule
        Training Topic Opinion & \textit{System Message} &
        You believe that \{\text{train}ing\_topic\_statement ($x_{\text{train}}$)\} is \{opinion\_response ($o_{\text{train}}$)\}. & 
        You believe that \{States with stricter gun control laws have fewer gun deaths per capita.\} is \{Probably True\}. \\
        \midrule
        Query Topic Opinion & \textit{System Message} &
        You believe that that \{\text{query}\_topic\_statement ($x_{\text{query}}$)\} is \{opinion\_response ($o_{\text{query}}$)\}. & 
        You believe that \{The global climate is rapidly growing warmer.\} is \{Certainly True\}. \\        
        \midrule
        Query & \textit{User Message} &
        Now, what is your opinion on the following statement using the following scale of responses?         
        \newline
        \newline
         \{\text{query}\_topic\_statement ($x_{\text{query}}$)\} is Certainly False, \{\text{query}\_topic\_statement ($x_{\text{query}}$)\} is Probably False, \{\text{query}\_topic\_statement ($x_{\text{query}}$)\} is Lean False, \{\text{query}\_topic\_statement ($x_{\text{query}}$)\} is Lean True, \{\text{query}\_topic\_statement ($x_{\text{query}}$)\} is Probably True, \{\text{query}\_topic\_statement ($x_{\text{query}}$)\} is Certainly True. 
        \newline
        \newline
        Statement: \{\text{query}\_topic\_statement ($x_{\text{query}}$)\} 
        \newline
        \newline
        Your opinion on the scale of responses:
        &
        
        Now, what is your opinion on the following statement using the following scale of responses? 
        \newline
        \newline        
        \{The global climate is rapidly growing warmer.\} is Certainly False, \{The global climate is rapidly growing warmer.\} is Probably False, \{The global climate is rapidly growing warmer.\} is Lean False,  \{The global climate is rapidly growing warmer., Probably True that \{The global climate is rapidly growing warmer.\} is Lean True, \{The global climate is rapidly growing warmer.\} is Certainly True
        \newline
        \newline
        Statement: \{The global climate is rapidly growing warmer.\}
        \newline
        \newline        
        Your opinion on the scale of responses:
        \\
        \bottomrule
    \end{tabular}
}
\caption{The prompts used for the LLM agent construction and querying in the in-context learning setting.}
\label{tab:list_prompt_icl}
\end{table*}

\section{The Prompts for LLM Agent Construction Through Supervised Fine-tuning (SFT)}
\label{app:list_prompt_sft}

Table~\ref{tab:list_prompt_sft} shows the prompts we use to construct and query the LLM agents in the supervised fine-tuning setting (\Scref{sec:llm_agent_construction}). The demographic information is included in the system message in the same prompt template as in \Scref{app:list_prompt_icl}. For the topic-specific opinions, however, instead of including them in the prompt, we formulate them as (prompt, response) pairs for supervised fine-tuning, where prompt is the input and response is the output. The prompt templates and examples are shown in Table~\ref{tab:list_prompt_sft}.

\begin{table*}[th!]
\small
\centering
\resizebox{\linewidth}{!}{  
    \begin{tabular}{@{}p{7cm}p{7cm}p{2cm}p{2cm}@{}}
        \toprule
        Prompt Template & Example Prompt & Response Template & Example Response\\
        \toprule
        What is your opinion on the following statement using the following scale of responses? 
        \newline
        \newline        
        Certainly False that \{query\_topic\_statement ($x_{\text{query}}$)\}, Probably False that \{query\_topic\_statement ($x_{\text{query}}$)\}, Maybe False that \{query\_topic\_statement ($x_{\text{query}}$)\}, Maybe True that \{query\_topic\_statement ($x_{\text{query}}$)\}, Probably True that \{query\_topic\_statement ($x_{\text{query}}$)\}, Certainly True that \{query\_topic\_statement ($x_{\text{query}}$)\} Statement: \{query\_topic\_statement ($x_{\text{query}}$)\}. 
        \newline
        \newline
        Please choose your response from the following list of options: Certainly False, Probably False, Maybe False, Maybe True, Probably True, Certainly True. &
        What is your opinion on the following statement using the following scale of responses? 
        \newline
        \newline        
        Certainly False that \{States with stricter gun control laws have fewer gun deaths per capita\}, Probably False that \{States with stricter gun control laws have fewer gun deaths per capita\}, Maybe False that \{States with stricter gun control laws have fewer gun deaths per capita\}, Maybe True that \{States with stricter gun control laws have fewer gun deaths per capita\}, Probably True that \{States with stricter gun control laws have fewer gun deaths per capita\}, Certainly True that \{States with stricter gun control laws have fewer gun deaths per capita\} Statement: \{States with stricter gun control laws have fewer gun deaths per capita\} 
        \newline
        \newline
        Please choose your response from the following list of options: Certainly False, Probably False, Maybe False, Maybe True, Probably True, Certainly True. & 
        My Response: \{opinion\_response\} & 
        My Response: \{Certainly True\} \\        
        \bottomrule
    \end{tabular}
}
\caption{The prompts used for the LLM agent construction and querying in the supervised fine-tuning setting.}
\label{tab:list_prompt_sft}
\end{table*}

\section{The Choice of Number of Factors in Factor Analysis}
\label{app:fa_number_factors}
\begin{figure}[th!] 
\centering
\includegraphics[width=1\linewidth]{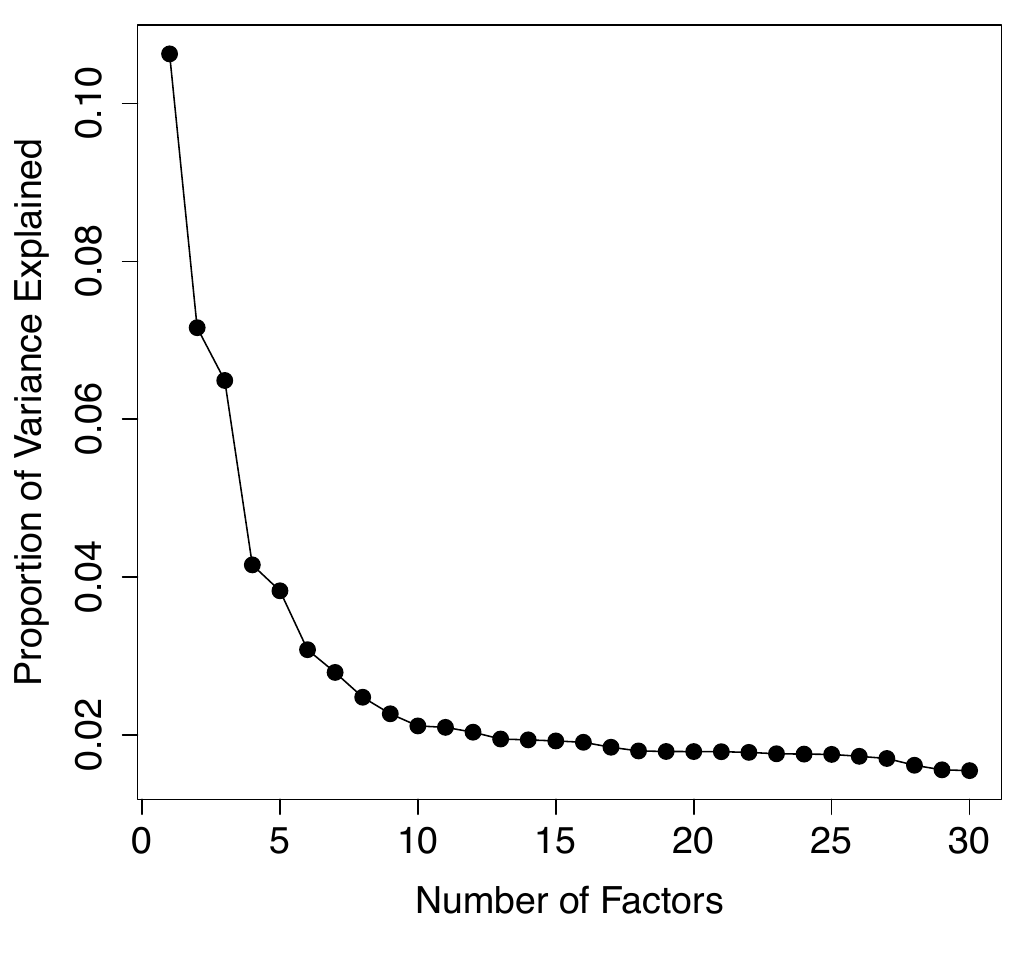}
\vspace{-2mm}
\caption{The scree plot of the factor analysis solution.} 
\label{fig:fa_scree_plot}
\vspace{-4mm}
\end{figure}

To determine the number of factors to retain in our factor analysis (FA), we visualize the scree plot in Figure~\ref{fig:fa_scree_plot}. We see that the explained variance plateaus after including 9 factors (the ``elbow point''). Therefore, we decide to retain 9 factors.
\section{Supervised Fine-tuning Details}
\label{app:fine_tuning}
In this section, we elaborate the different strategies used for constructing LLM agents through supervised fine-tuning.
\vspace{-3mm}
\begin{enumerate}[label=\alph*.,leftmargin=15pt]\itemsep-0.3em
    \item \textbf{Baseline: No-Demo.} Baseline without fine-tuning, (identical to same condition in ICL.
    \item \textbf{Baseline: Demo.} Baseline without fine-tuning, identical to same condition in ICL.
    \item \textbf{Demo+Train [same category]:} For each topic category we constructed the dataset $\Dcal_{\text{SFT}}=\{(d_i,x_{\text{train},i}),o_{\text{train},i}\}_{i=1}^N$. We then fine-tuned the LLM with input context providing the demographic information along with the training topic statement $(d,x_{\text{train}})$, and using the corresponding human Likert-scale response $o_{\text{train}}$ as the target. After fine-tuning, we assessed the LLM agent's opinion on query topics $x_{\text{query}}$ belonging to the same topic category $x_{\text{train}}$ \footnote{For example, we fine-tuned an LLM on the respondents' opinions on the training topic for the Ghost category, then queried its opinion on the test topics in the Ghost category.}. This is the critical condition of interest that tests cross-topic generalization. The verbatim prompts are in \Scref{app:list_prompt_sft}.
    \item \textbf{Baseline: Demo+Train [random category].}: Similar to Demo+Train [same category] condition, but the training topic opinion ($x_{\text{train}}^{\dagger}$, $o_{\text{train}}^{\dagger}$) is from a different topic category as the query topic $\xquery$, allowing us to assess whether generalization is restricted to topics in the same belief category.
    \item \textbf{Upper Bound: Demo+Train+Query.} Upper bound without fine-tuning, identical to same condition in ICL.   
\end{enumerate}


ChatGPT (\texttt{gpt-3.5-turbo-0125}) is fine-tuned through OpenAI's fine-tuning API \footnote{\url{https://platform.openai.com/docs/guides/fine-tuning}}. These were the hyper-parameters used in fine-tuning:
\begin{itemize}
    \item{Number of Epochs:} 3
    \item{Batch Size:} 1
    \item{Learning Rate Multiplier:} 2
\end{itemize}
\section{The Full Factor Analysis Results}
\label{app:full_factor_loading_matrix}
\begin{figure*}[th!] 
\centering
\includegraphics[width=0.9\linewidth]{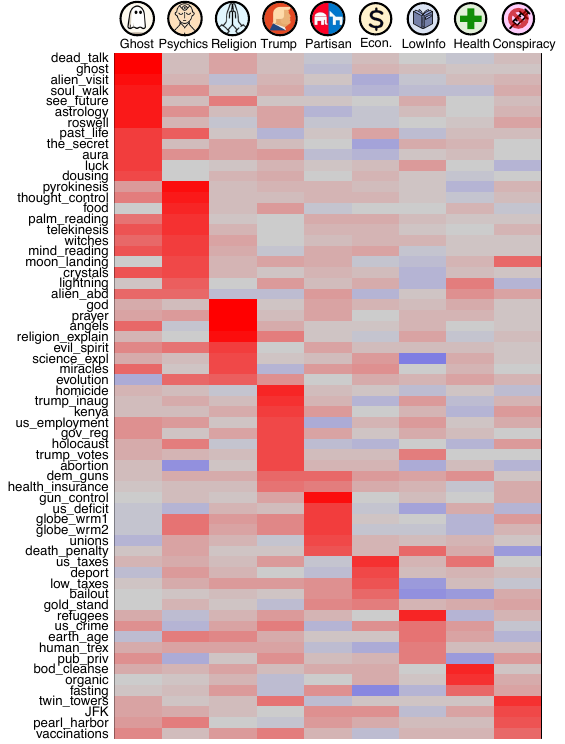}
\vspace{-2mm}
\caption{The factor loading matrix of the Controversial Belief Survey. The column indicates the nine factor, and the rows are the 64 topics. Red indicates topics that load highly on a factor, gray indicates near 0 loading, and blue indicates loading in the negative direction.  We focus on the Ghost category and Partisan categories, highlighted by the green box and the violet box respectively. The topics in the Ghost category has minimal loading on the Partisan factor and vice versa (highlighted by the black boxes). The full statement of each topic is in Table~\ref{tab:list_topic} (\Scref{app:list_topic}).} 
\label{fig:fa_loading_matrix}
\vspace{-4mm}
\end{figure*}

In Figure~\ref{fig:belief_network_all_factors}b in the main text, we only show the factor loading matrix of the Ghost and the Partisan factors, and the corresponding topics. In this section, we discuss the full factor analysis result. 

The factor analysis reveals nine latent factors underlying the 64 topics. Figure~\ref{fig:fa_loading_matrix} shows the full factor loading matrix. The red blocks highlight strong correlations among opinions within each factor, indicating that endorsing one conception in a cluster often predicts opinion in other conceptions within the same cluster. We assign the name of each factor based on its constituent topics: Ghost, Psychics, Religion, Trump, Partisan, Economic, LowInfo, Health, and Conspiracy. The 64 topics are categorized by which factor they have the  highest loadings on. For instance, the topic about communication with the dead belongs to the Ghost category because it has the highest loading on the Ghost factor (Table~\ref{tab:list_topic} shows the full list of topics and categories).
\section{Compute Resources}
\label{app:compute_resources}
We ran all experiments with Mistral and LLaMA 3.1 on a GPU machine
equipped with 1x NVIDIA A100. The experiments with ChatGPT and GPT-4-o-mini cost about 400 USD.
\section{Sensitivity Analysis}
\label{app:sensitivity}
\begin{table}[tb!]
\centering
\small
\resizebox{\linewidth}{!}{
    \begin{tabular}{@{}lrrr@{}}
        \toprule
        Condition & \multicolumn{3}{c}{Temperature} \\
        \cmidrule(l){2-4} 
        & 0 & 0.7 & 1 \\
        \midrule
        \textit{Baselines} \\
        \hspace{1em} No-Demo & 1.80 & 1.68 & 1.66 \\
        \hspace{1em} Demo & 1.70 & 1.70 & 1.71 \\
        \hspace{1em} Demo + Train [Rand. Cat.] & 1.66 & 1.67 & 1.68 \\
        \hspace{1em} Train [Same Cat.] & 1.43 & 1.48 & 1.49 \\
        \textbf{Demo + Train [Same Cat.]} & \textbf{1.37} & \textbf{1.34} & \textbf{1.44} \\
        \textit{Upper bound} \\
        \hspace{1em} Demo + Same Train + Truth & 0.42 & 0.42 & 0.53 \\
        \midrule        
        Average Relative Gain (\%) & 18.57 & 22.54 & 14.64 \\
        \bottomrule
    \end{tabular}
}
\caption{Average $\mae$ and average relative gain of each LLM agent (powering by ChatGPT) construction condition across three temperature values.}
\label{tab:result_llm_vs_human_chatgpt_sensitivity_temperature}
\end{table}

\paragraph{Sensitivity Analyses} We evaluate the sensitivity of our result to randomness due to different temperature values when using temperature sampling. Across $T \in \{0, 0.7, 1\}$ using ChatGPT, the results show consistent trends (Table~\ref{tab:result_llm_vs_human_chatgpt_sensitivity_temperature}).

\end{document}